%% file: nips_2018.tex
\documentclass{article}

\pdfoutput=1

\usepackage[numbers]{natbib}

\usepackage[nonatbib, preprint]{nips_2018}

\usepackage[utf8]{inputenc} 
\usepackage[T1]{fontenc}    
\usepackage{hyperref}       
\usepackage{url}            
\usepackage{booktabs}       
\usepackage{amsfonts}       
\usepackage{nicefrac}       
\usepackage{microtype}      

\usepackage{subcaption}
\usepackage{graphicx}

\title{Lidar Cloud Detection with Fully Convolutional Networks}

\author{
   Erol Cromwell \\
  Pacific Northwest National Laboratory\\
  Richland, WA 99352 \\
  \texttt{erol.cromwell@pnnl.gov} \\
  \And
   Donna Flynn \\
  Pacific Northwest National Laboratory\\
  Richland, WA 99352 \\
  \texttt{donna.flynn@pnnl.gov} \\
}

\begin{document}

\maketitle

\input{paper_structure}

\end{document}

%% file: paper_structure.tex
\begin{abstract}

\input{abstract}

\end{abstract}

\section{Introduction} \label{introduction}

\input{introduction}

\section{Dataset} \label{mpl_data}

\input{mpl_data}

\section{Related work} \label{related_work}
\input{related_work}

\section{Data preprocessing and preparation} \label{dataset}
\input{dataset}

\section{Model architecture} \label{model}
\input{model}

\section{Training} \label{train}
\input{training}

\section{Results} \label{results}
\input{results}

\section{Conclusion} \label{conclusion}
\input{conclusion}

\subsection*{Acknowledgements}

%

\input{acknowledgements}


\bibliographystyle{plain}
\bibliography{nips_2018}

%% file: abstract.tex
In this contribution, 
we present a novel approach for 
segmenting laser radar (lidar) imagery into geometric time-height cloud locations
with a fully convolutional network (FCN).
We describe a semi-supervised learning method 
to train the FCN by: 
pre-training the classification layers of the FCN with image-level annotations,
pre-training the entire FCN with the cloud locations of the MPLCMASK cloud mask algorithm, 
and fully supervised learning with hand-labeled cloud locations. 
We show the model achieves higher levels of cloud identification compared to the 
cloud mask algorithm implementation.

%% file: introduction.tex
The vertical distribution of clouds from active remote sensing instrumentation is a widely 
used data product from global atmospheric measuring sites. The presence of clouds can be 
expressed as a time versus vertical height binary cloud mask (cloud 1, no cloud 0) and 
is a primary input for climate modeling efforts and cloud formation studies. 
Ground-based lidars have routinely been employed to provide these cloud masks. 
Current cloud detection algorithms used to retrieve these masks from lidar data, however, 
do not always accurately identify the cloud boundaries and tend to oversample or 
over-represent the cloud. This translates as uncertainty for assessing the radiative 
impact of clouds and tracking changes in cloud climatologies. 
Additionally, such algorithms require significant effort to develop and maintain and 
are sensitive to instrument changes and accurate instrument calibration.


Machine learning has only recently been applied to the atmospheric science domain and in particular 
cloud retrievals from lidar data. 
Previous applications of machine learning to lidar data have been primarily focused on airborne 
lidar instruments for tasks such as estimating vegetation height and canopy cover \citep{Stojanova2010} 
and estimating forest biomass \citep{Gleason2012}. 
Recently, a neural network was trained to detect clouds from satellite data,
but in a limited setting with supervised learning \citep{Gomez2017}. 
However, deep learning techniques have not been used for cloud detection from ground-based lidar instruments.

In our work, we present a method for using FCNs to detect clouds from MPL data
that surpasses an implementation of the well-established 
Wang and Sasson cloud mask algorithm \citep{Wang2001}.
We develop a semi-supervised learning method to train the FCN,
involving pre-training the ``classification'' weights of the model
with image-level annotations
and pre-training the entire model with noisy annotations.
We demonstrate the trained FCN model performs better than the established algorithm and 
verify our learning methodology improves the model performance compared to only using 
end-to-end supervised learning.

The remainder of this paper is organized as follows.
In section \ref{mpl_data}, we introduce the lidar system and data used for this study. 
In section \ref{related_work}, 
we discuss some related work on using FCNs 
for image segmenation and semi-supervised learning methods.
Section \ref{dataset} outlines the preprocessing of our training and test data
and sections \ref{model} and \ref{train} describe our model architecture and training methodology. 
Section \ref{results} describes and analyzes our cloud segmentation 
results.
We conclude with some suggestions for future work in section \ref{conclusion}.

%% file: mpl_data.tex
For this study we use lidar data from micropulse lidar (MPL) systems deployed by the the 
Atmospheric Radiation Measurement (ARM).
The MPL system operate by emitting a low energy laser pulse (\textasciitilde 10 $\mu$J) combined with high 
pulse rate (2500 Hz). 
Pulses are emitted in an alternating linearly and circularly polarized state and each 
return signal that is scattered back is collected using a photon detector. 
The return signal from each pulse is time-resolved to provide the vertical height of 
the aerosol particles or cloud droplets that are responsible for the scattering. 
These return signals measured by the photon detector are processed and combined to 
then provide vertical profiles of the atmospheric structure as attenuated backscatter. 
Additionally, the linear depolarization ratio (LDR) can be calculated from the linear and 
circularly return signals to help determine cloud phase (liquid, ice or mixed) as 
well distinguishing aerosol and clear air.  
A detailed description of the MPL data processing is given by Campbell et al. \citep{Campbell2002} and 
the calculation of the LDR is provided in Flynn et al. \citep{Flynn2007}.

The MPL data stream used is sgp30smplcmask1zwangC1 from the Southern Great Plains (SGP) 
C1 facility located near Lamont Oklahoma \citep{mplmask}. 
This location is selected because this observatory is the world’s largest and most extensive 
climate research facility and focal point for modeling and forecasting efforts. 
We selected data from January 2010 to December 2016 to train the model. 
One 30smplcmask1zwang data file contains a 24-hour period of lidar profiles at 
30 second temporal resolution and 30 m vertical resolution out to 18km (667 x 2880 data points). 
We use the total attenuated backscatter, linear depolarization ratio (LDR) as well as a 
cloud mask product using an algorithm based on Wang and Sassen \citep{Wang2001}. 
The implemented algorithm relies in part on the strong change in the slope of the 
backscatter that is observed in the presence of clouds and distinguishing clouds 
from noise or an aerosol layer by inspecting the ratio of extinction of the lidar 
signal to the backscatter \citep{sivaraman2011}. 
This cloud mask product, MPLCMASK, is used to compare the model’s performance and pre-train the model. 

%% file: related_work.tex

An FCN is a neural network that contains convolutional layers, but
no fully-connected layers.
Long et al. \citep{Long_2015_CVPR} show how FCNs can be used to  
accurately segment images semantically with pixel-to-pixel predictions.
The advantage of FCNs is they can be trained with end-to-end,
pixel-to-pixel operations which increase learning and inference 
and allow for fusion of finer and coarser features to
help performance \citep{Long_2015_CVPR}.
Several applications of FCNs in scientific domains include:
biomedical imaging for cells \citep{UNET}, 
brain tumor segmentation \citep{Havaei2017},
uranium particle identification from mass spectrometry imagery \citep{Tarolli2017},
and mitigating radio frequency interference signals in radio data \citep{akeret2017radio}.
We use an image-segmentation approach with an FCN model for several reasons.
First, it is easy to translate our problem, identify clouds from MPL data,
to an image segmentation problem.
The input data can be treated as an image
by having each measurement act as a color channel. 
Then, our ``image'' is a two-channel image, with the backscatter as one channel
and the LDR as the other.
Then, the problem becomes segmenting the input image for
regions of clouds and as previously stated,
FCN's have been very successful
performing pixel-by-pixel image segmentation.
Visual interpretation of cloud features in lidar images is very similar to a segmentation 
approach, which takes advantage of global and local features within an image thus 
suggesting FCNs are an appropriate and natural choice.

An advancement in deep convolutional networks for image segmentation 
has been the use of weakly-supervised and semi-supervised learning techniques
to train the networks.
Hong et al. \citep{Hong2015} proposes a deep neural network consisting of a
classification network, a segmentation network, and bridging layers connecting
the two networks. 
By separating the network, they train each part of the network independently,
training the classification network with image-level annotations 
and the segmentation network with full image semantic segmentations. 
Their semi-supervised approach is competitive against other
semi- and weakly-supervised trained models.
Papandreou et al. \citep{Papandreou2015}
uses weakly (bounding-box and image-level) annotated images to 
train deep convolutional neural networks.
They show that training with a small amount of pixel-level annotated images
combined with a large amount of weakly annotated images
results in performance very close to that of training on only 
pixel-level annotations.
Similarly, 
Weihman et al. \citep{Weihman2016} demonstrated using end-to-end unsupervised learning as a
pre-training step on FCNs produces models that perform statistically similar
to their fully supervised counterparts.
By adding a reconstruction output layer, their model is able to 
use the reconstruction loss as weighted component of the loss function
and slowly transition to the segmentation loss during training.
The advantage of these techniques is  
models can be trained using large amounts of unlabeled and weakly labeled data
while requiring minimal ground truth data.
This is applicable to our work 
since manually create a binary cloud mask for lidar data 
is a time and labor intensive process,
resulting in a small dataset of hand-labeled data.
Leveraging these training methodologies removes the need for large datasets of hand-labeled cloud masks
without sacrificing model performance.

%% file: dataset.tex
\input{example_back_ldr}

To improve the visualization of the lidar imagery we take the log of the attenuated backscatter.
Any missing values, infinite values, or not-a-number (NaN) values are then set to the daily minimum value. 
Each day is then zero-centered by subtracting the mean from the backscatter and normalized by 
dividing by the standard deviation of the backscatter mean. The LDR valid data range is between 0 and 1. 
Missing values and NaNs are set to 0, any values greater than 1 are set to 1 and values 
less than 0 are set to 0. 
Figure \ref{fig:example_back_ldr} shows an example of the backscatter and LDR images.

To train and test the model, 82 days of ground truth are hand labeled by a lidar expert with 
extensive experience with the MPL data. 
This ground truth consists of cloud masks that have been labeled using the attenuated 
backscatter and LDR measurements to visually identify the cloud locations. 
There are 54 days (30 from January 2015, 24 from February 2015) are used for training and 
validation and 28 (March 2015) are held out for testing.

Given the small amount of hand-labeled data, 
we increase the amount of data eightfold by splitting
each day into quarters time-wise with some overlap (667 x 800)
and adding the horizontally flipped version of each quarter day to the dataset.
The data was split so that each quarter mostly identified
a unique part of the profile and to have max-pooling
operations reduce the time dimension more evenly.
The quarter days are determined by: 
time bins 0-800,
time bins 680-1480, 
time bins 1400-2200, 
and time bins 2080-2880.

We divide the data into four separate disjoint datasets for training and testing:

\textbf{Classification dataset} \hspace{0.1cm} For pre-training the classification part of the model,
a set of 1780 quarter days is used with
image-level annotations identifying whether the quarter days contains clouds.
A quarter day is marked as containing clouds if the MPLCMASK contains clouds.
It is acceptable to use the MPLCMASK to identify if the quarter day has cloud 
since the main issue with the product is that it oversamples the cloud boundaries,
not misidentify whether clouds exist at a given time.
Thus, the product can still capture whether there are clouds in the image.
The dataset is divided evenly between quarter days with clouds (890)
and without clouds (890).

\textbf{Noisy dataset} \hspace{0.1cm} For pre-training the entire model,
a set of 4200 quarter days is used (4000 with clouds, 200 without clouds).
The ground truth data for this set is the MPLCMASK product. 
We treat the MPLCMASK product as noisy annotations, since the mask oversamples the shape of
the clouds and merges cloud boundaries.
No hand-labeled days used for training are included in this dataset.

\textbf{Hand-labeled dataset} \hspace{0.1cm} The dataset for supervised training and fine-tuning the model consists of
432 quarter days (362 with cloud, 70 without).
This consist of the January and February 2015 lidar data.
The ground truth data for this set is the hand-labeled data.

\textbf{Holdout dataset} \hspace{0.1cm} Finally, the March 2015 hand-labeled data is used as a hold-out dataset to test 
the full segmentation model results.
This consists of 224 quarter days (172 with cloud, 52 without).

%% file: example_back_ldr.tex
\begin{figure*}[h]
\centering
    \begin{subfigure}[b]{0.40\textwidth}
    \centering
        \includegraphics[scale=0.12]{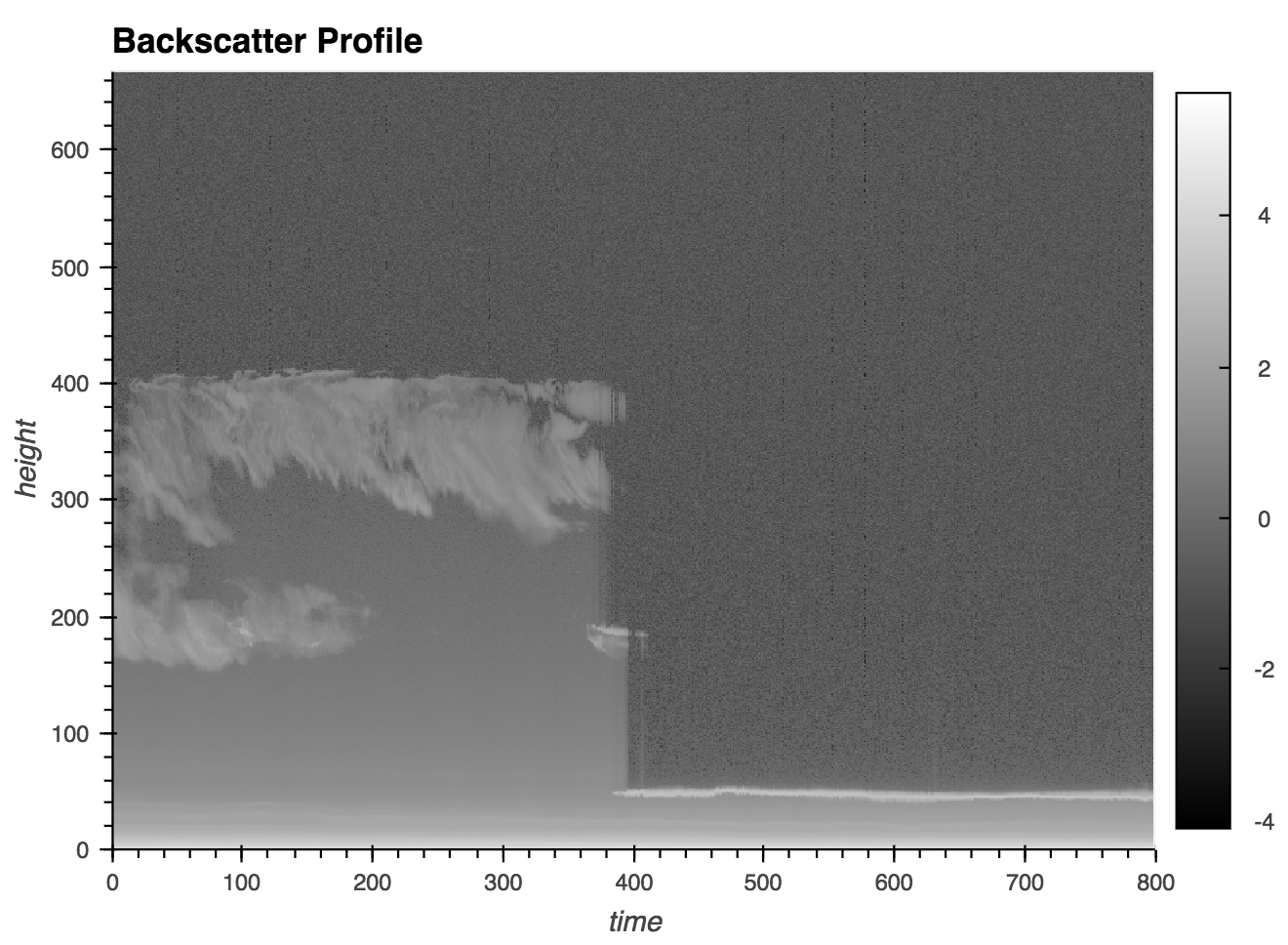}
    \end{subfigure}
    \begin{subfigure}[b]{0.40\textwidth}
    \centering
        \includegraphics[scale=0.12]{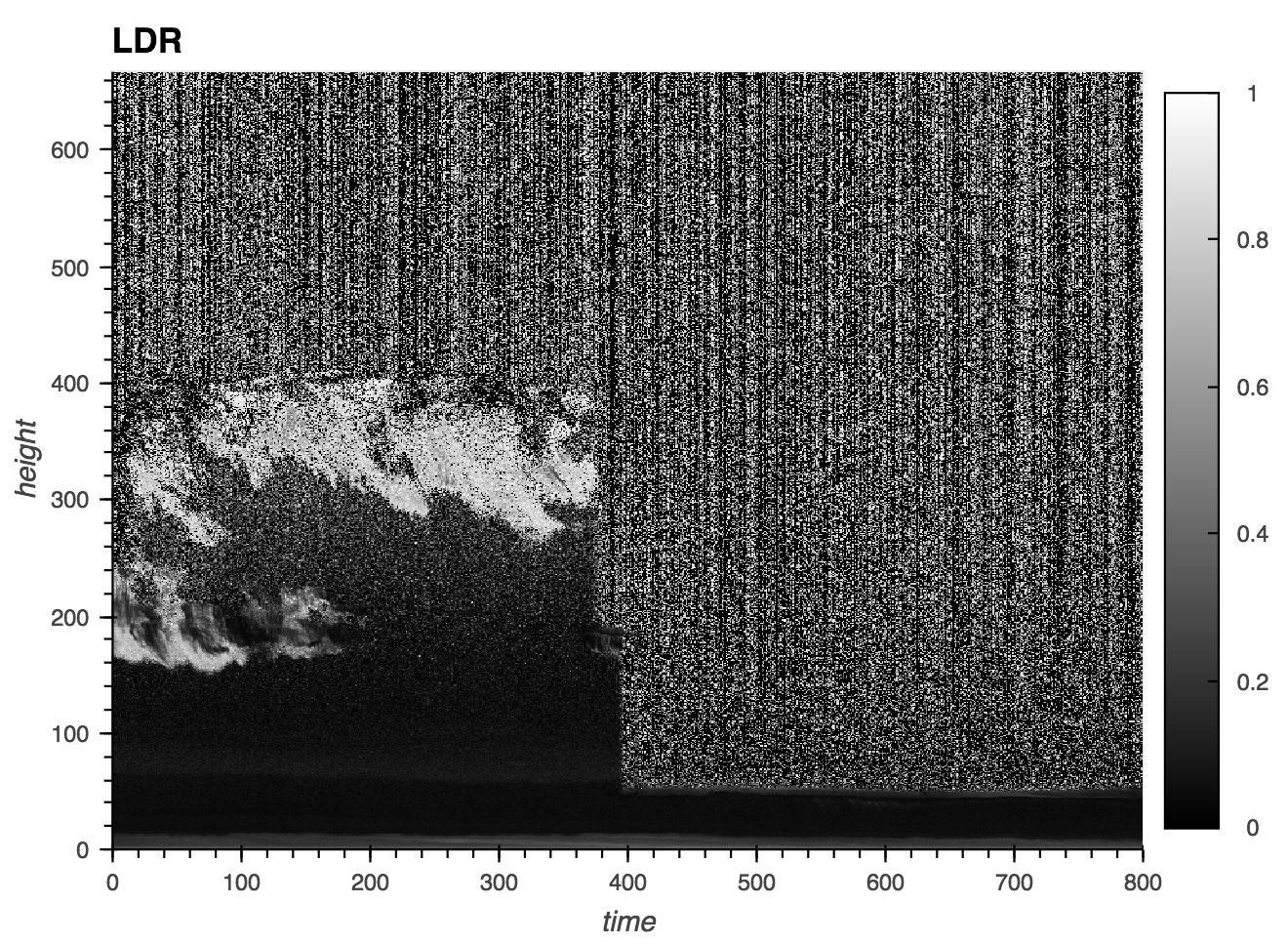}
    \end{subfigure}
\caption{Example of MPL backscatter profile (left) and 
         LDR (right). From January 1st, 2015 (time bins 0-880).}
\label{fig:example_back_ldr}
\end{figure*}

%% file: model.tex
\begin{figure*}[ht]
\centering
\includegraphics[width=1.0\linewidth]{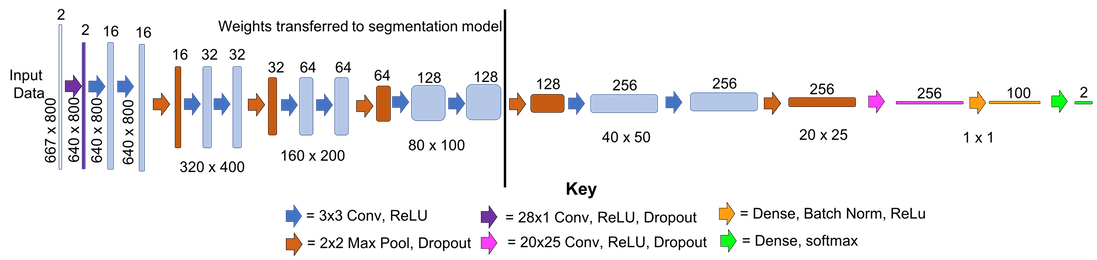}
\caption{Diagram of classification neural network trained to identify if lidar data has a cloud}
\label{classify_model}
\end{figure*}

We use the U-net model design \citep{UNET} as our base FCN model architecture.
The U-net design consists of downsampling the input image then upscaling back to the original size
in a series of convolutional layers and deconvolutional layers.
In the same spirit of the Hong et al. decoupled network \citep{Hong2015}, 
the model is divided into two parts during training: 
the classification part of the model and the segmentation part.

\subsection{Classification model}

The model input is the quarter day backscatter profiles and LDR data (667 x 800 x 2).
The first layer is a 28 x 1 convolution with stride of 1 that reduces the data size to 640 x 800 x 2.
The initial size reduction is done to ensure the downsampling (and later on upsampling)
dimension changes are consistent throughout the entire model.
Next are five convolutional-pooling blocks.
Each block contains
two 3 x 3 convolutions layers with stride of 1 with 
rectified linear unit (ReLU) activation function
and a 2 x 2 max pooling layer with dropout.
The first convolutional layer doubles the depth size in each block
except for the first one, 
which increase the depth size from 2 to 16.
This is followed by a 20 x 25 convolutional layer with stride of 1 with ReLU and dropout
to flatten the layer.
Next are two dense layers, the first using batch normalization and ReLU 
and the second using a softmax activation function to reduce to a 1 x 2 output.
The first index in the output array is the probability the input data does not contain a cloud
and the seconds is the probability the input data contains a cloud.
Figure \ref{classify_model} shows the classification model.

\subsection{Segmentation model}

\begin{figure*}[h]
\centering
\includegraphics[width=0.7\linewidth]{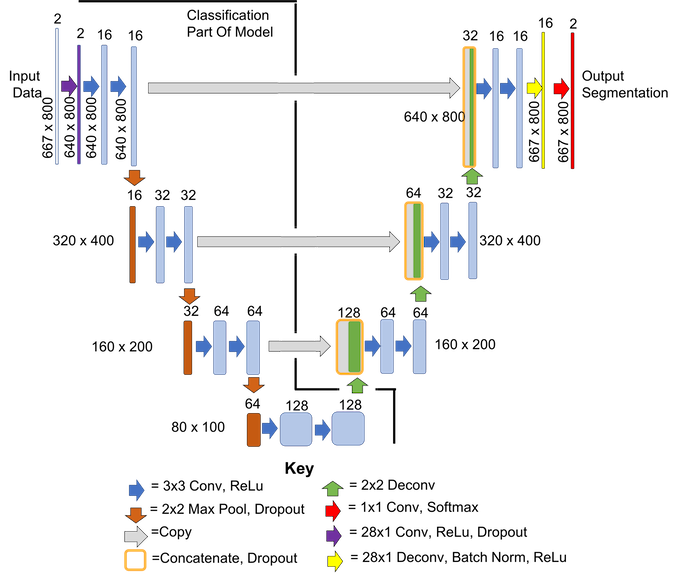}
\caption{Diagram of fully convolutional neural network trained to identify clouds from lidar data}
\label{seg_model}
\end{figure*}

The segmentation model architecture uses the same structure as the classification model
up to and including the last 80 x 100 x 128 convolutional output
(excluding the max-pooling and dropout layer).
The convolutional output is connected to three deconvolutional blocks. 
Each deconvolutional block consists of a deconvolutional layer that upsamples 
the image size by 2 and reduces the number of filters by a factor of 2.  
This is concatenated with the output of the last convolutional layer of 
the same image size as the output of the deconvolution layer,
followed by two 3 x 3 convolutional layers. 
These "skip" connections help transfer higher level features from earlier convolutions
later in model training.
We use a 28 x 1 deconvolution layer with batch normalization and ReLU activation
to return the data to the initial time and height dimensions (667 x 800).
Finally, the model performs a 1 x 1 convolutional layer with softmax activation function
to create the output layer.
The output layer is the softmax probability for each time-height point whether it is a cloud or not a cloud.
A point is identified as being a cloud if it has a probability greater than or equal to 50\%
of being a cloud.
Figure \ref{seg_model} shows the full segmentation model and notes the layers 
from the classification model that are reused.

%% file: training.tex
The FCN model is trained in three stages.
First, the classification model in figure \ref{classify_model} 
is trained using the image-level annotations from the classification dataset.
Then, we transfer the classification weights from the model 
(the weights from the layers on the "classification" side in figure \ref{seg_model})
to the segmentation model.
Second,
we pre-train the FCN segmentation model with the noisy annotations of the noisy dataset.
This is similar to using unsupervised pre-training,
since it involves training the model on data that is not the desired output
to improve the overall performance.
However, unsupervised pre-training tends to involve
reconstructing the input data for the output
while the noisy annotations (the MPLCMASK product) 
are semantically related to the desired output (the hand-labeled cloud mask).
By transferring the weights of the classification model and training on the noisy annotations, 
we reduce the segmentation search space of the entire model
to make it easier to train the model with a limited sample of full segmentation labels.
Finally, 
the FCN segmentation model is trained and fine-tuned using the hand-labeled dataset
with the hand-labeled cloud mask as ground truth.

Training for the classification weights and pre-training the segmentation model
is done over 100 epochs
and fine-tuning the segmentation model is done over 300 epochs.
We did not deem it necessary to train the classification model or 
pre-train the entire model for longer than 100 epochs,
since we are transferring the weights from the training steps
and
the end goal is not the best cloud classification model or 
a mimic of the MPLCMASK product.
We use a batch size of 32 for pre-training the classification model 
and 10 for pre-training and training the segmentation model. 
We use accuracy for the loss function of the classification model 
and categorical cross entropy for the loss function of the segmentation model.
Each training step uses an Adam optimizer \cite{Kingma2014} 
and a 70-15-15 train-val-test split of the dataset.

\begin{table}[htp]
\caption{Best hyper-parameters for each model training}
\begin{center}
\begin{tabular}{|c|c|c|c|}
\hline
Training step & learning rate & dropout & decay rate \\
\hline
1) Image-level annotation & 0.001 & 0.5 & 0.0 \\
\hline
2) Noisy annotation & 0.00059 & 0.2 & 0.0 \\
\hline
3) Hand-labeled annotations & 0.001 & 0.0 & 0.0 \\
\hline

\end{tabular}
\end{center}
\label{table:hyper}
\end{table}%

In each training phase, we perform a hyperparameter search
over the learning rate, dropout, and decay rate. 
For each set of models in the search, 
we keep the weights from the epoch where the model has the best validation score 
(accuracy for classification model, F1-score for segmentation model).
Then, we select the model with the highest validation score.
For the final training step, the selected model is the one with best
F1-score on the hold-out dataset. 
Table \ref{table:hyper} lists the best parameters for each training step.

For the FCN segmentation model, we use 
precision (\% of predicted clouds that are actual clouds),
recall (\% of clouds that are predicted as clouds),
and F1-score (harmonic mean of the precision and recall) 
as our performance metrics.
Each of the metrics are calculated over the entire dataset, and not per input image. 
We use F1-score for the segmentation model instead of accuracy
since only a very small set of the time-height points in the hand-labeled ground truth are labeled as clouds.
In the hold-out dataset, only \%5.79 of points are clouds
and in the hand-labeled dataset, only \%4.65 of points are clouds.
As such, the accuracy measurement (sum of the true positive clouds and true negative non-clouds
divided by the total number of points)
gets skewed extremely high because the number of non-cloud points 
heavily outweighs the number of cloud points.
For example, if there are 100 points of which 90 are non-cloud and 10 are cloud and
the model correctly identifies 90 non-cloud and 1 cloud, then the accuracy is 91\%
despite only correctly identifying 10\% of the cloud points.

The model is implemented with the Keras python module \cite{chollet2015keras}.
We train the model on a computer cluster with
Dual Intel Broadwell E5-2620 v4 @ 2.10GHz CPUs node with
64GB 2133Mhz DDR4 memory and
dual NVIDIA P100 12GB PCI-e based GPUs.

%% file: results.tex
\input{test_results_table}
\input{march_results_table}
Overall, the FCN segmentation model outperforms
the MPLCMASK product.
For the test split of the hand-labeled dataset (table \ref{table:test_results})
and the holdout dataset (table \ref{table:march_results}),
the FCN model has an F1-score of 0.8777 and 0.8508, respectively.
This exceeds the performance of the MPLCMASK (0.5892 and 0.65, respectively).
The FCN model precision is almost double,
indicating the model correctly identifies more clouds than
the MPLCMASK product by a factor of 2.
Thus, the model is able to captures more of the cloud detail in the output
than the MPLCMASK.
We note that the FCN model slightly underperforms against the algorithm
in recall in the holdout dataset (0.8687 and 0.9049, respectively),
but outperforms the algorithm in the test split (0.8998 and 0.89, respectively).

%
%

\input{good_figures}

Figure \ref{fig:good_data} presents several qualitative results from the FCN model.
As shown, the clouds identified by the model closely follow that
of the hand-labeled mask.
In the first and third examples (March 13th and March 26th), 
the model output is more detailed 
in contrast to the MPLCMASK product,
which tends to exaggerate the cloud shape and size and merge multiple cloud layers.
In the second example (March 16th), the MPLCMASK
is unable to consistently detect the cloud layer as indicated
by the vertical gaps in the mask.



\subsection{Training methodology verification}
To verify the training methodology,
we train a FCN model without the first two training steps (i.e., no image-level and noisy annotation pre-training)
and one without the second training step (no noisy annotation pre-training).
For the second model, we use the same classification weights that were transferred to the best FCN model
during the first training step.
Additionally, we use the same hyperparameter search techniques 
to find the best model.
As seen in table \ref{table:march_results} (rows 2 and 3), 
both of these models have lower F1-scores (0.8263 and 0.8242) and 
precision (0.7801 and 0.7938) than the fully trained FCN model on the hold-out dataset.
We do note the FCN model with no pre-training does have a slightly higher recall (0.8783)
than the fully trained model.
On the test-split of hand-labeled dataset, the fully trained FCN model outperforms both of them 
in each category (see rows 2 and 3 of table \ref{table:test_results}).
Thus, the pre-training with the image-level and noisy annotated data increases the model's overall performance.

%
%

\input{bad_march_figures}

\subsection{Misidentified aerosol-dust layers}

In the holdout dataset (March 2015), we discovered one day, last quarter of March 30th 2015, 
where the model performs very poorly (f1-score: 0.2929, precision: 0.1806, recall: 0.7752). 
The model incorrectly marks a layer in the lower height bins (see figure \ref{fig:march_30_data}) as cloud. 
The attenuated backscatter of the scatterers in this layer visually appears similar to 
aerosol but the LDR values are close to that of an ice leading the model to label the 
layer cloud. 
This layer is not observed in images from a ceilometer, a second co-located lidar system 
at the SGP, on this same day. 
The ceilometer has a longer wavelength and is less sensitive to detecting smaller 
particles such as aerosol. 
This suggest this layer is likely aerosol, possibly related to local agricultural activity.
Looking closer at the training data, there are no days in January or February 2015 that 
had similar aerosol layers. Thus, the model did not have the opportunity to train on these 
edge cases and did not learn to ignore this layer.

\input{oliktok_figures}

\subsection{Test on MPL data from different facility}
In addition, we test the FCN model on MPL data
from the ARM Oliktok mobile facility on the north slope of Alaska 
to see how well the model performs on data from a different site location \citep{mplmask}
since different sites have different atmospheric properties and weather
that can affect the input data.
We randomly select 14 days from May 2016 to hand-label and run the model on.
The model performs reasonably well on the Oliktok data, 
achieving an F1-score of 0.8045 (0.8478 precision, 0.7654 recall).
This performs much better than the MPLCMASK product, 
which has an F1-score of 0.4185 (0.371 precision, 0.48 recall).
Several qualitative results can be seen in figure \ref{fig:oliktok_figures}.
While the FCN model does perform noticeably worse on the Oliktok data in contrast to the
holdout data from the SGP site, the results are encouraging.
We believe if we use transfer learning and train the model on data from the Oliktok site,
we can improve the model's performance on the Oliktok site data and its overall performance.

%% file: test_results_table.tex
\begin{table*}[ht]
\caption{Model performance on test split of hand-labeled dataset}
\centering
\begin{tabular}{|c|c|c|c|}
\multicolumn{1}{c}{\bf Method} & \multicolumn{1}{c}{\bf F1-Score} & \multicolumn{1}{c}{\bf Precision} & \multicolumn{1}{c}{\bf Recall} \\
\hline
MPLCMASK product & 0.5892 & 0.4423 & 0.8795 \\
\hline
No pre-training & 0.792 & 0.7301 & 0.8653 \\
\hline
No noisy annotation pre-training & 0.8126 & 0.7795 & 0.8487 \\ 
\hline
FCN segmentation model & 0.8777 & 0.8637 & 0.8998 \\
\hline
\end{tabular}
\label{table:test_results}
\end{table*}

%% file: march_results_table.tex
\begin{table*}[ht]
\caption{Model performance on holdout dataset (March 2015)}
\centering
\begin{tabular}{|c|c|c|c|}
\multicolumn{1}{c}{\bf Method} & \multicolumn{1}{c}{\bf F1-Score} & \multicolumn{1}{c}{\bf Precision} & \multicolumn{1}{c}{\bf Recall} \\
\hline
MPLCMASK product & 0.65 & 0.5072 & 0.9049 \\
\hline
No pre-training & 0.8263 & 0.7801 & 0.8783 \\
\hline
No noisy annotation pre-training & 0.8242 & 0.7938 & 0.857 \\
\hline
FCN segmentation model & 0.8508 & 0.8336 & 0.8687 \\
\hline
\end{tabular}
\label{table:march_results}
\end{table*}

%% file: good_figures.tex
\begin{figure}[t]
\centering
    \begin{subfigure}[b]{0.19\linewidth}
    \centering
        \caption{backscatter}
        \includegraphics[scale=0.115]{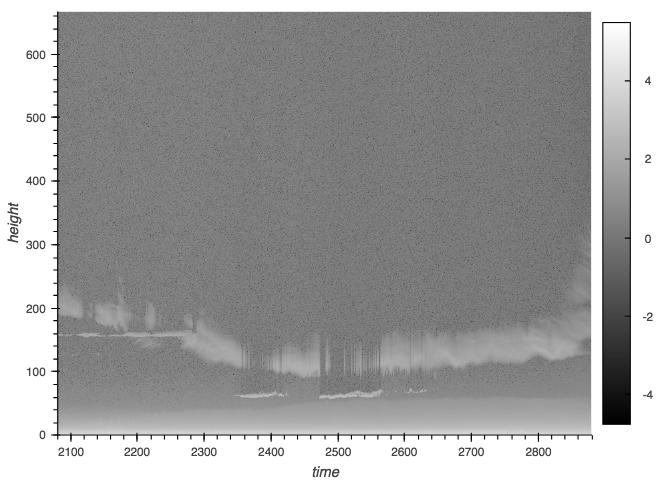}
    \end{subfigure}
    \begin{subfigure}[b]{0.19\linewidth}
    \centering
        \caption{ldr}
        \includegraphics[scale=0.115]{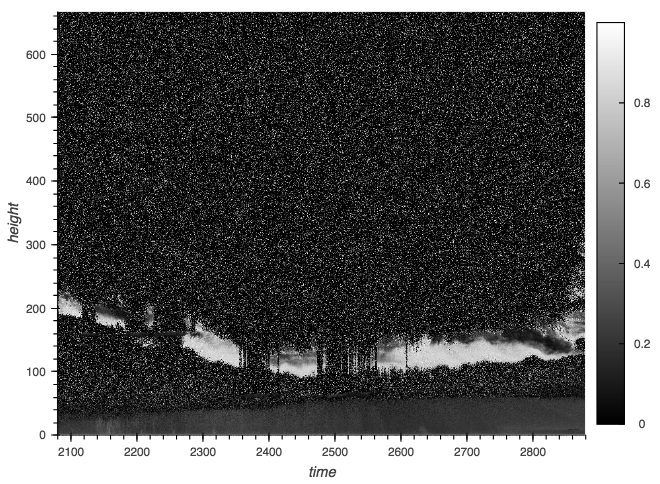}
    \end{subfigure}
    \begin{subfigure}[b]{0.19\linewidth}
    \centering
        \caption{hand-labeled}
        \includegraphics[scale=0.115]{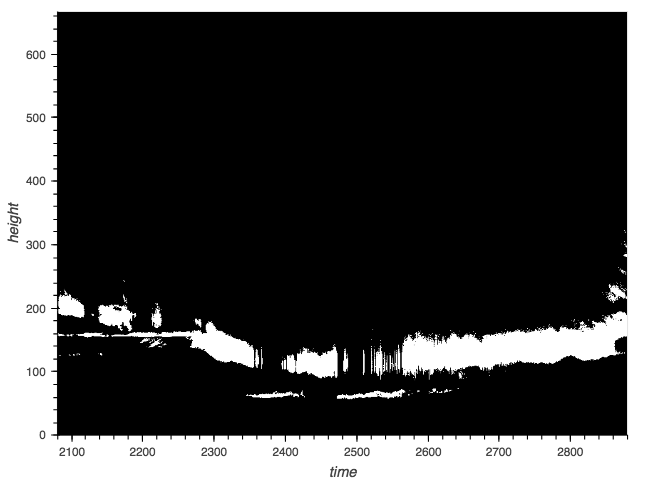}
    \end{subfigure}
    \begin{subfigure}[b]{0.19\linewidth}
    \centering
        \caption{model}
        \includegraphics[scale=0.115]{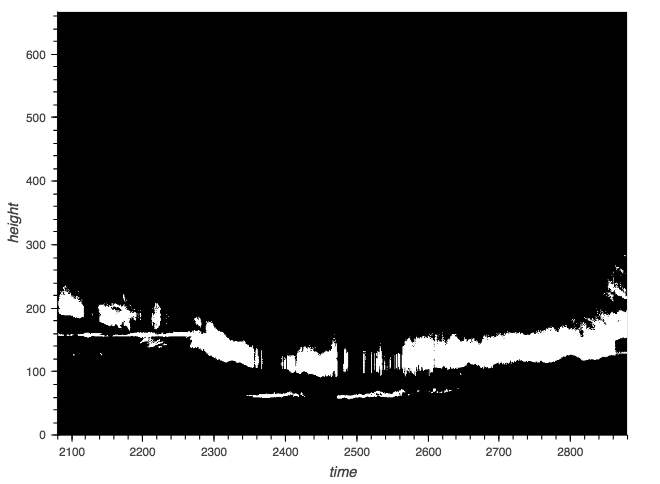}
    \end{subfigure}
    \begin{subfigure}[b]{0.19\linewidth}
    \centering
        \caption{MPLCMASK}
        \includegraphics[scale=0.115]{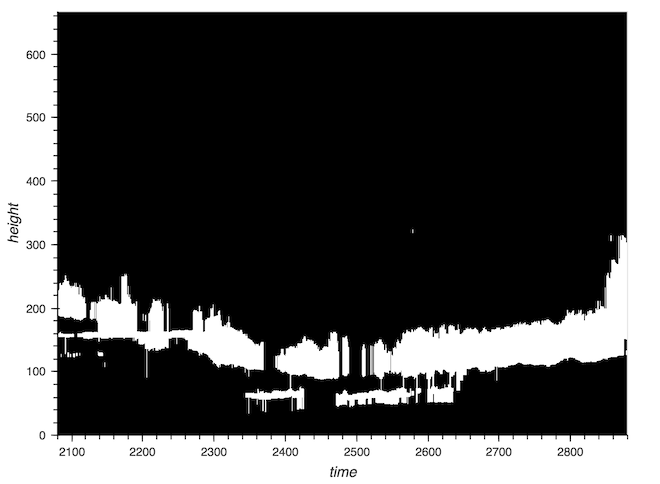}
    \end{subfigure}

    \begin{subfigure}[b]{0.19\linewidth}
    \centering
        \includegraphics[scale=0.115]{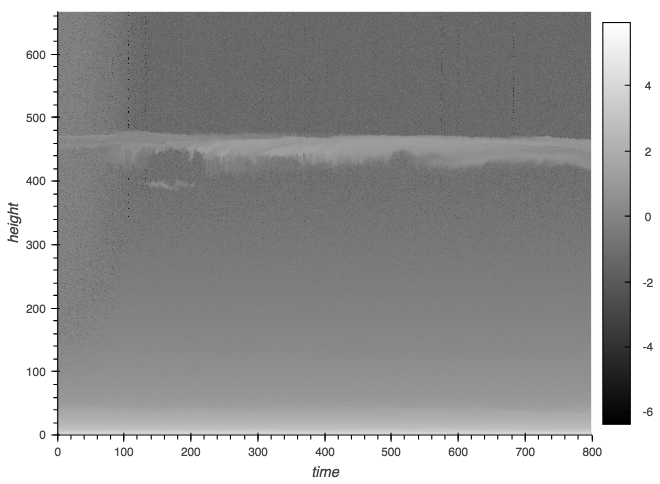}
    \end{subfigure}
    \begin{subfigure}[b]{0.19\linewidth}
    \centering
        \includegraphics[scale=0.115]{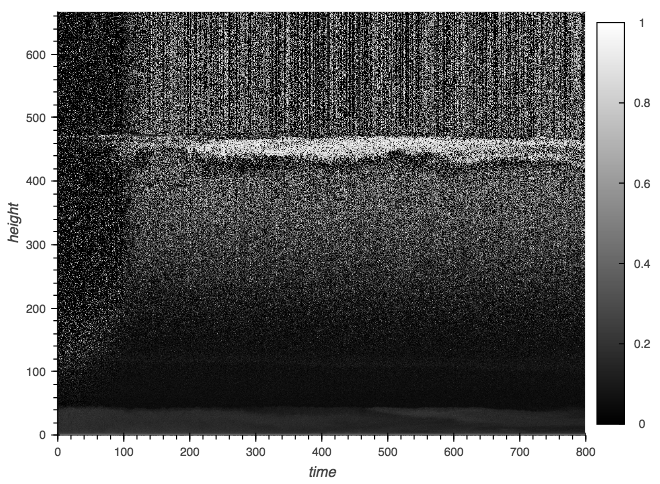}
    \end{subfigure}
    \begin{subfigure}[b]{0.19\linewidth}
    \centering
        \includegraphics[scale=0.115]{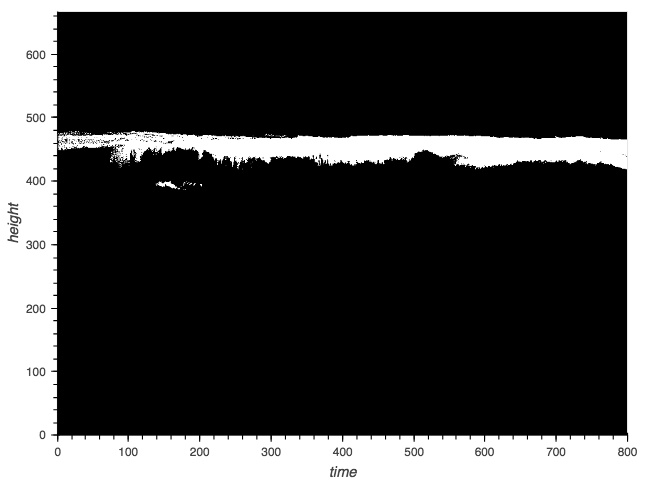}
    \end{subfigure}
    \begin{subfigure}[b]{0.19\linewidth}
    \centering
        \includegraphics[scale=0.115]{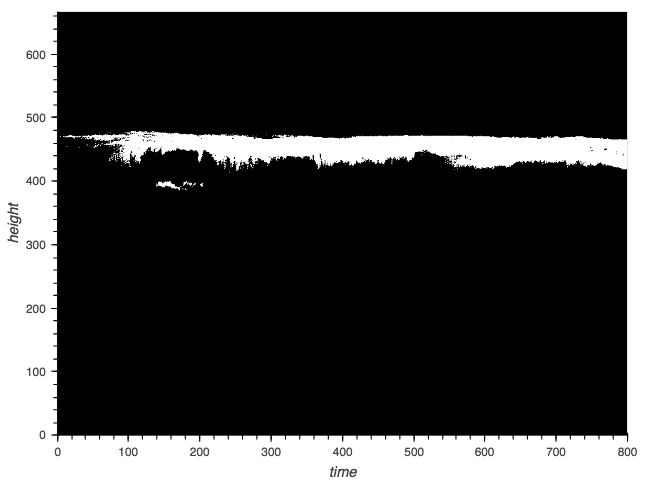}
    \end{subfigure}
    \begin{subfigure}[b]{0.19\linewidth}
    \centering
        \includegraphics[scale=0.115]{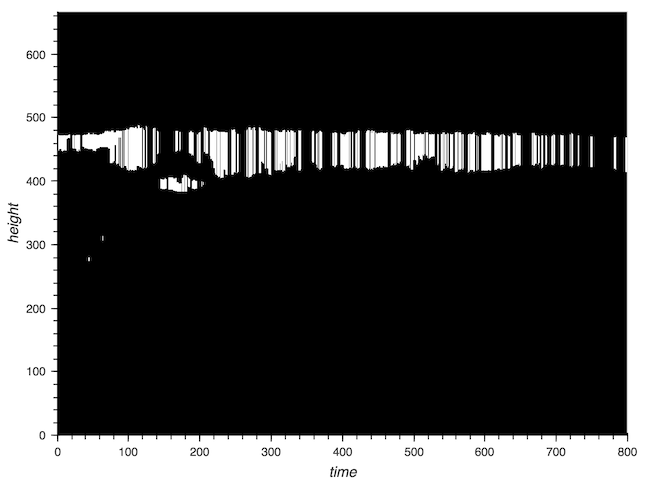}
    \end{subfigure}

    \begin{subfigure}[b]{0.19\linewidth}
    \centering
        \includegraphics[scale=0.115]{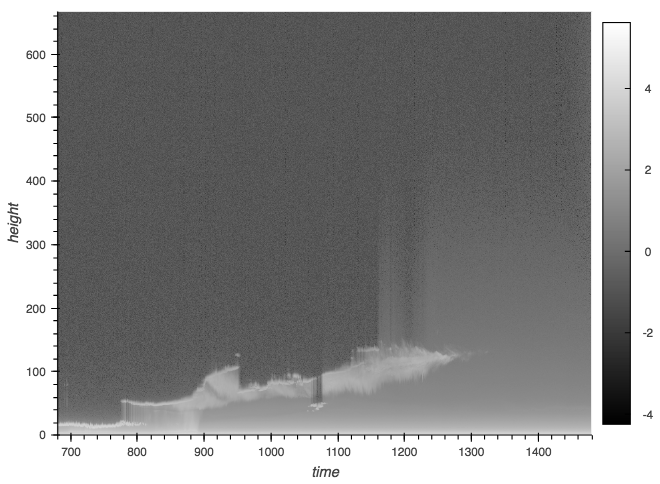}
    \end{subfigure}
    \begin{subfigure}[b]{0.19\linewidth}
    \centering
        \includegraphics[scale=0.115]{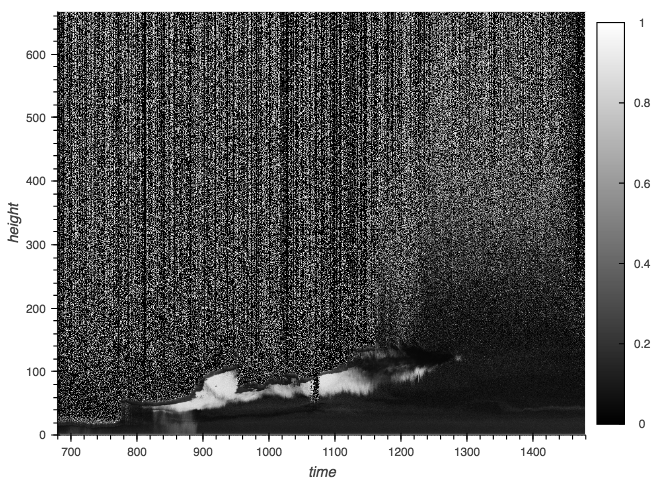}
    \end{subfigure}
    \begin{subfigure}[b]{0.19\linewidth}
    \centering
        \includegraphics[scale=0.115]{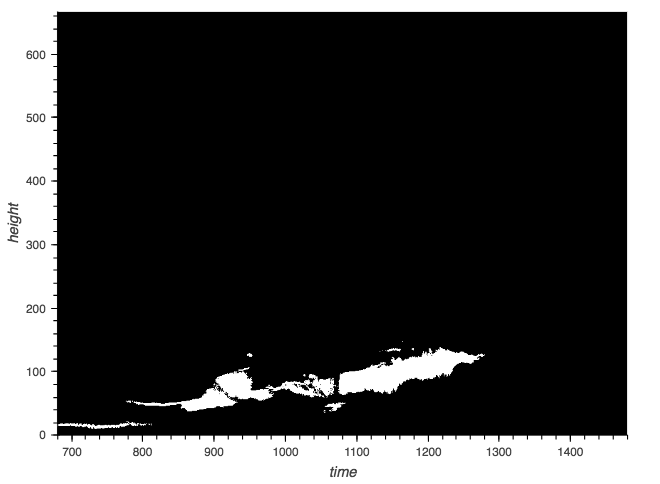}
    \end{subfigure}
    \begin{subfigure}[b]{0.19\linewidth}
    \centering
        \includegraphics[scale=0.115]{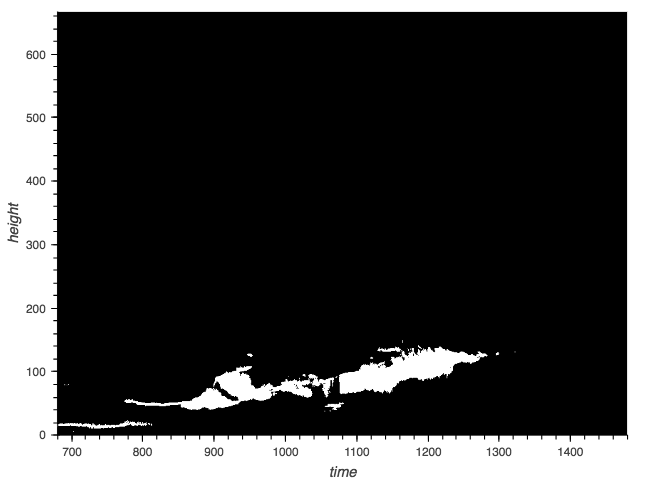}
    \end{subfigure}
    \begin{subfigure}[b]{0.19\linewidth}
    \centering
        \includegraphics[scale=0.115]{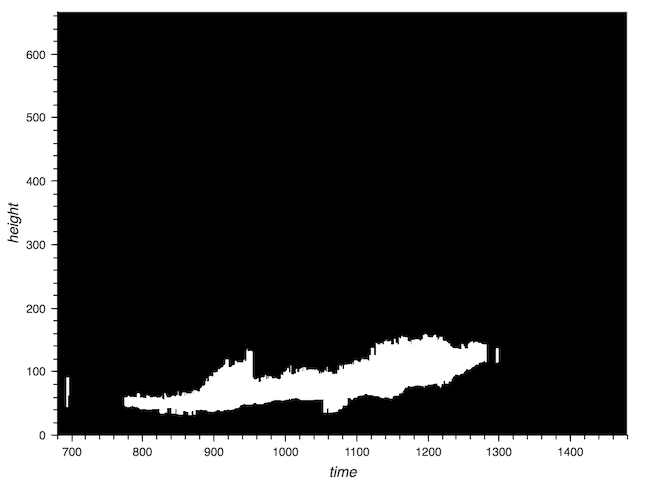}
    \end{subfigure}
\caption{Cloud segmentation results for several days 
         (top to bottom: March 13th, March 16th, March 26th, 2015).
         \textbf{(a)}: MPL backscatter profile.
         \textbf{(b)}: MPL linear depolarization ratio.
         \textbf{(c)}: hand-labeled cloud mask.
         \textbf{(d)}: segmentation model output.
         \textbf{(e)}: MPLCMASK cloud mask.}
\label{fig:good_data}
\end{figure}

%% file: bad_march_figures.tex
\begin{figure}[t]
\centering
    \begin{subfigure}[b]{0.19\linewidth}
    \centering
        \caption{backscatter}
        \includegraphics[scale=0.0575]{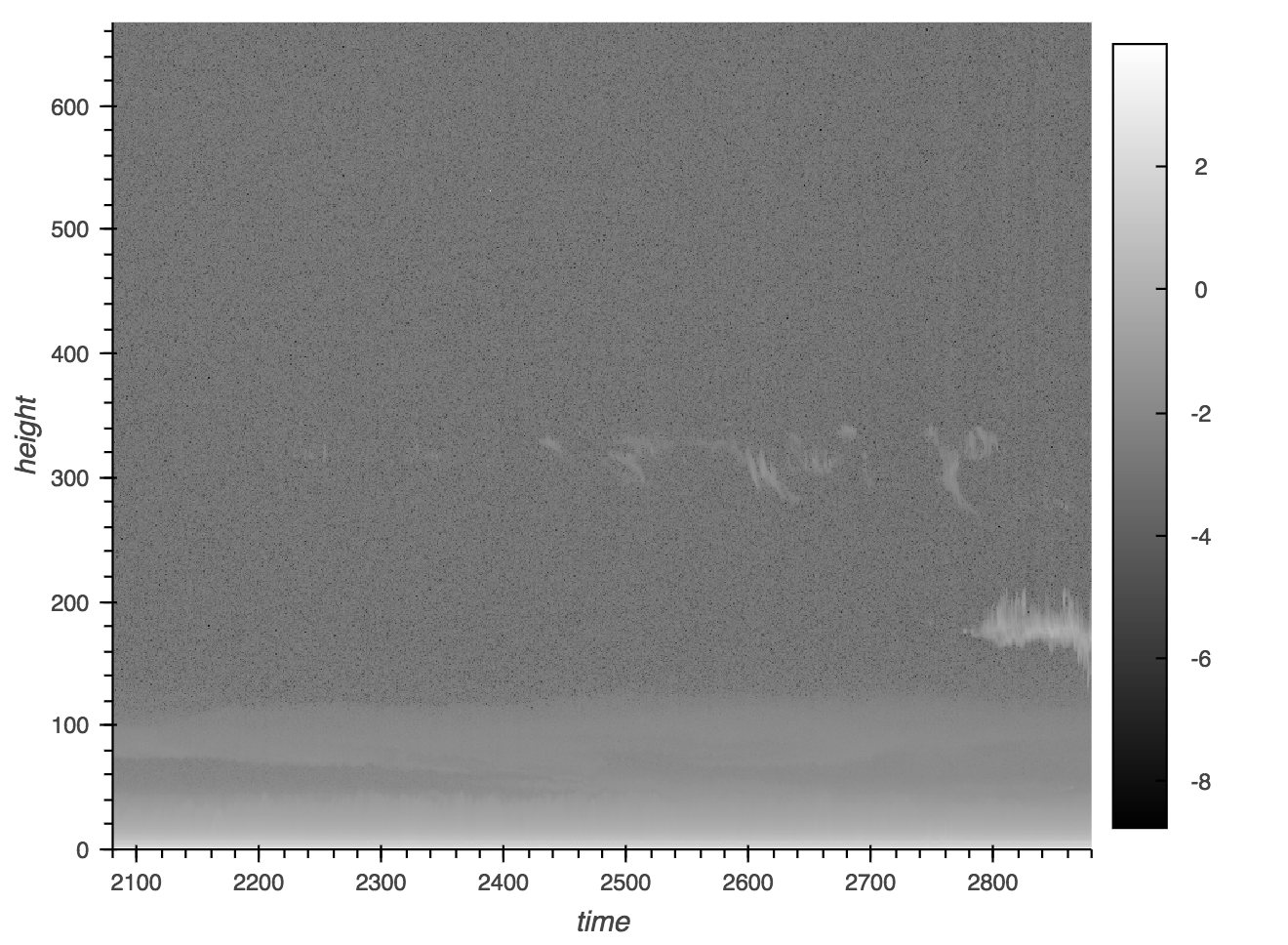}
    \end{subfigure}
    \begin{subfigure}[b]{0.19\linewidth}
    \centering
        \caption{ldr}
        \includegraphics[scale=0.0575]{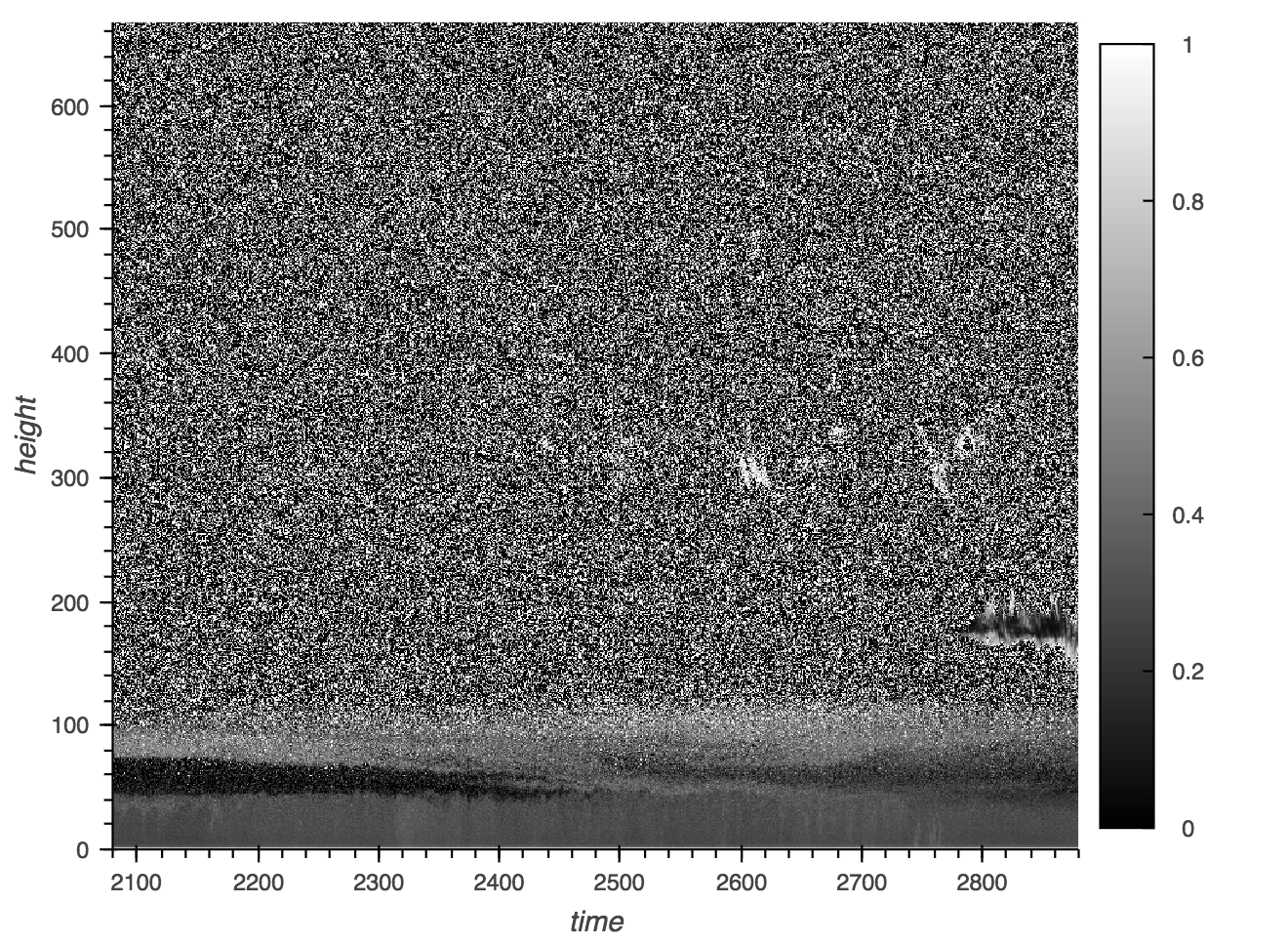}
    \end{subfigure}
    \begin{subfigure}[b]{0.19\linewidth}
    \centering
        \caption{hand-labeled}
        \includegraphics[scale=0.0575]{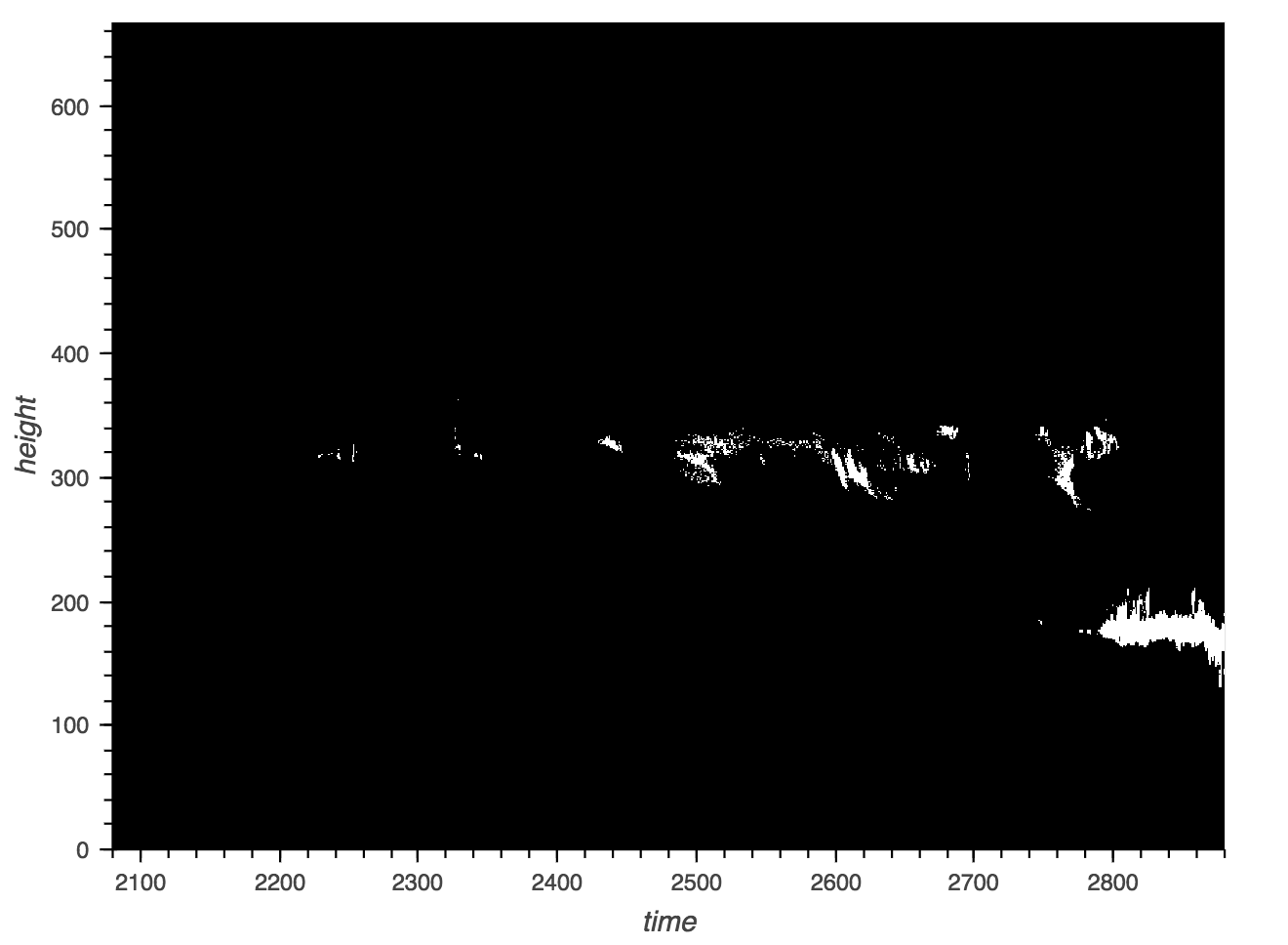}
    \end{subfigure}
    \begin{subfigure}[b]{0.19\linewidth}
    \centering
        \caption{model}
        \includegraphics[scale=0.0575]{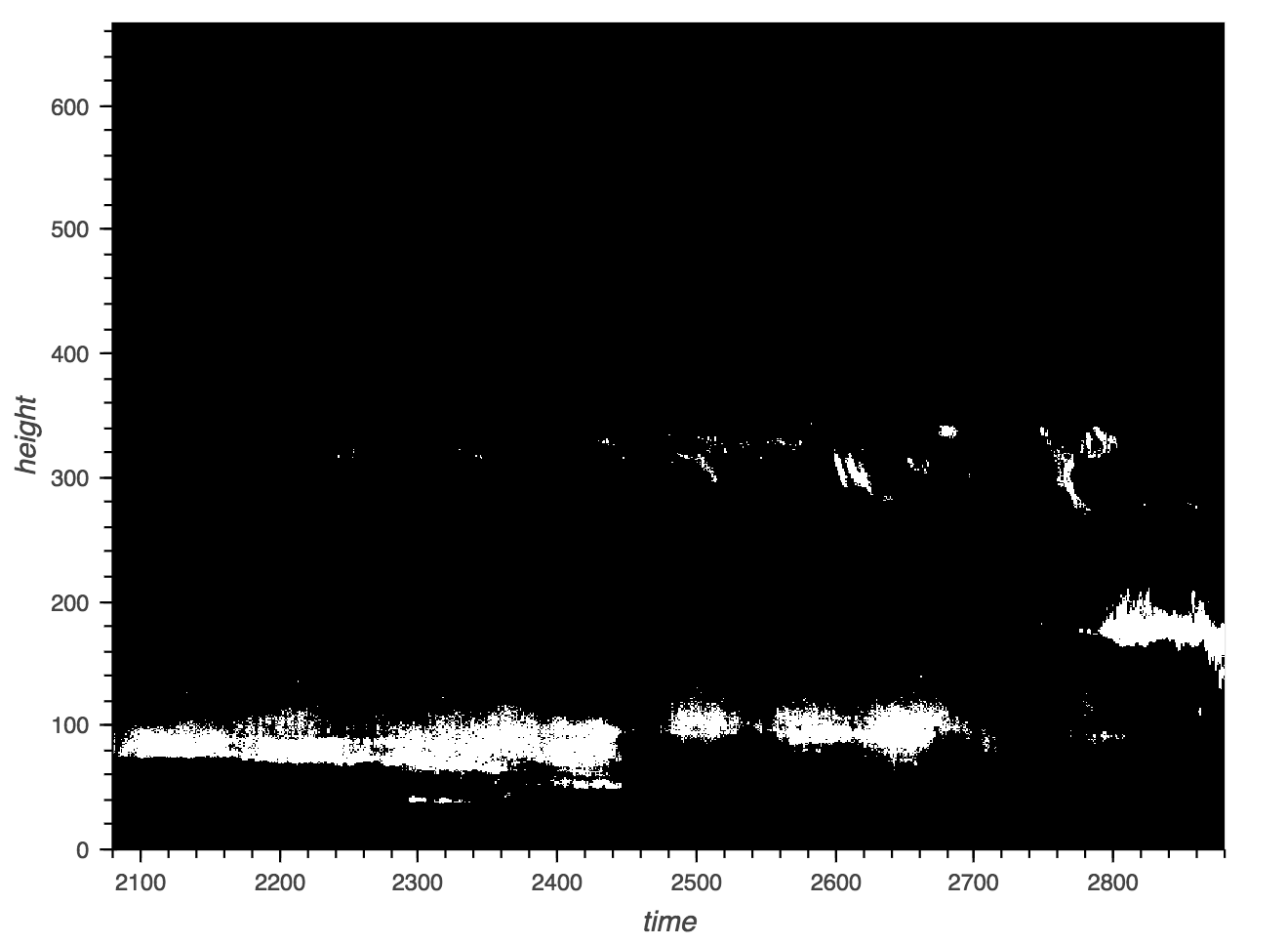}
    \end{subfigure}
    \begin{subfigure}[b]{0.19\linewidth}
    \centering
        \caption{MPLCMASK}
        \includegraphics[scale=0.0575]{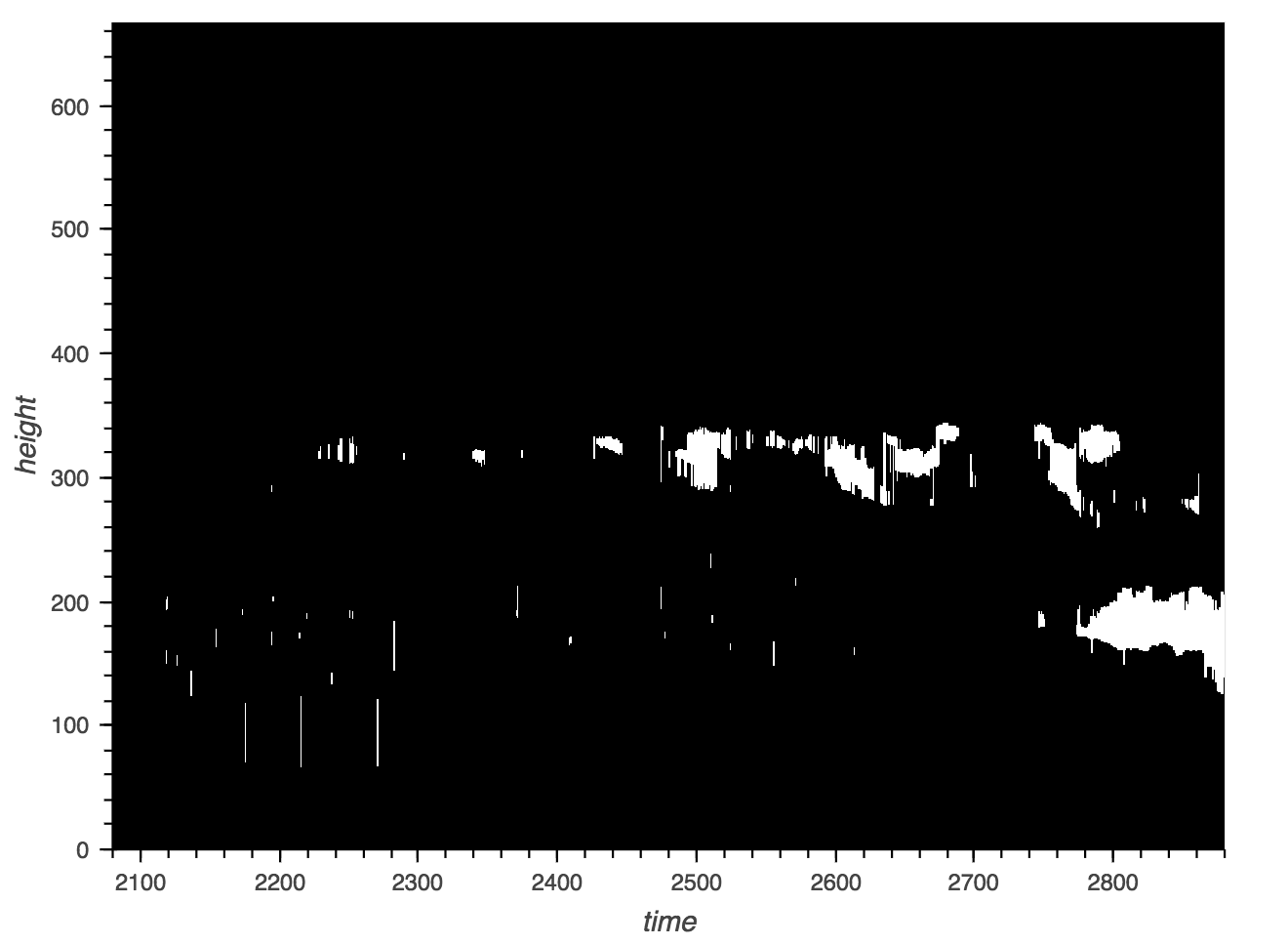}
    \end{subfigure}
\caption{Cloud segmentation results for March 30th, 2015.
         \textbf{(a)}: MPL backscatter profile.
         \textbf{(b)}: MPL linear depolarization ratio.
         \textbf{(c)}: hand-labeled cloud mask.
         \textbf{(d)}: segmentation model output.
         \textbf{(e)}: MPLCMASK cloud mask.}
\label{fig:march_30_data}
\end{figure}

%% file: oliktok_figures.tex
\begin{figure*}[t]
\centering
    \begin{subfigure}[b]{0.19\textwidth}
    \centering
        \caption{backscatter}
        \includegraphics[scale=0.118]{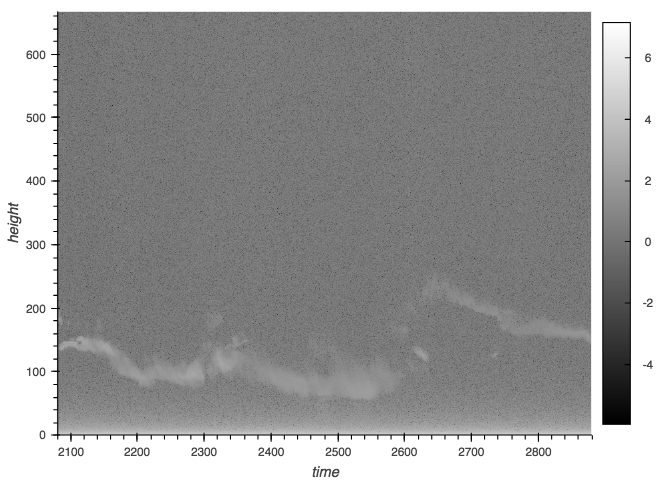}
    \end{subfigure}
    \begin{subfigure}[b]{0.19\textwidth}
    \centering
        \caption{ldr}
        \includegraphics[scale=0.118]{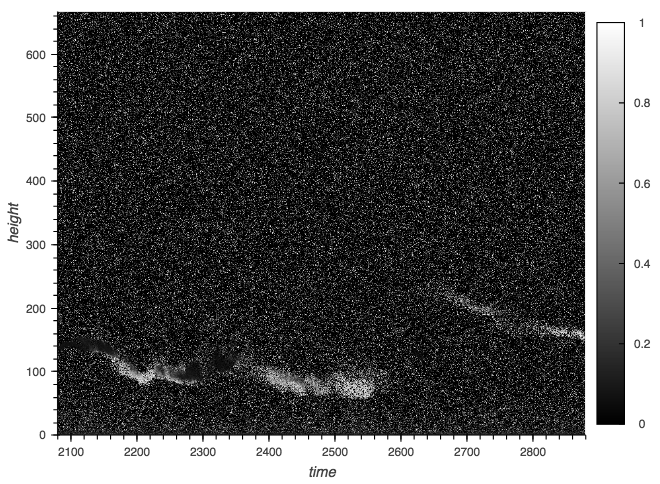}
    \end{subfigure}
    \begin{subfigure}[b]{0.19\textwidth}
    \centering
        \caption{hand-labeled}
        \includegraphics[scale=0.118]{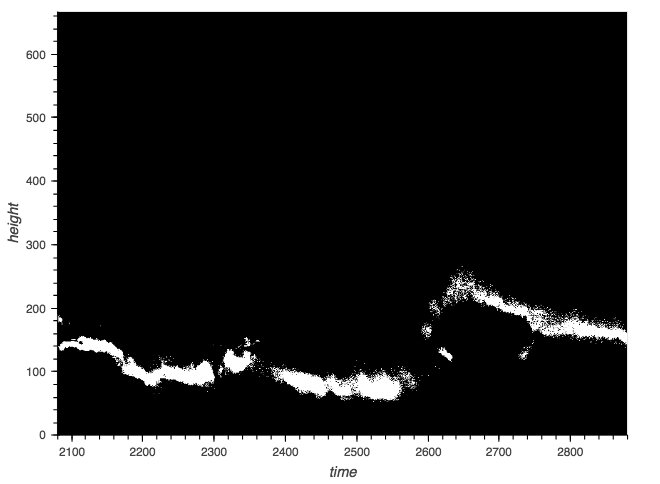}
    \end{subfigure}
    \begin{subfigure}[b]{0.19\textwidth}
    \centering
        \caption{model}
        \includegraphics[scale=0.118]{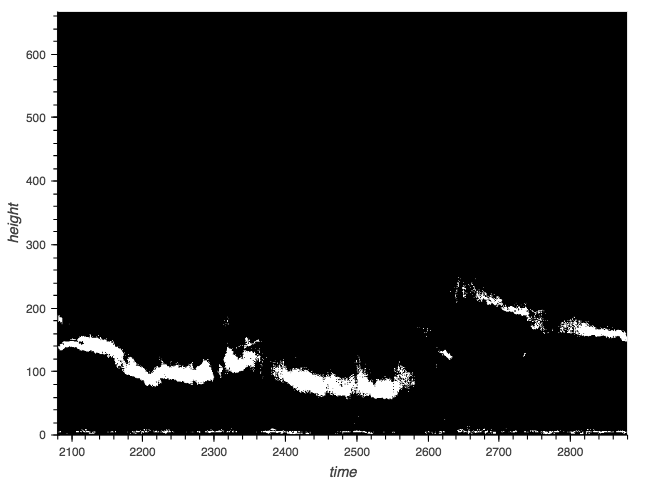}
    \end{subfigure}
    \begin{subfigure}[b]{0.19\textwidth}
    \centering
        \caption{MPLCMASK}
        \includegraphics[scale=0.118]{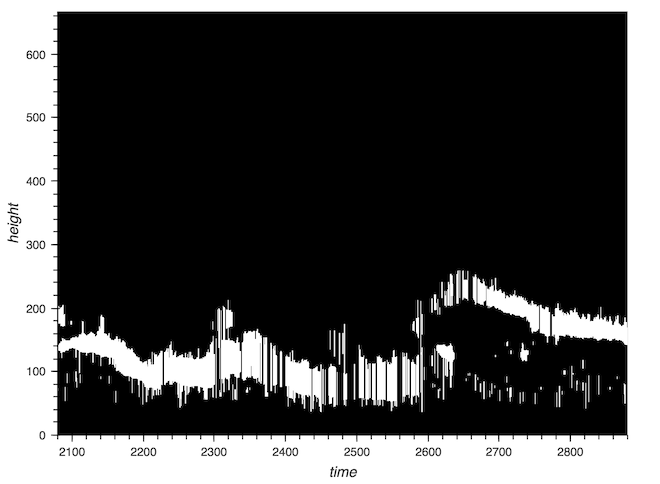}
    \end{subfigure}
    \newline
    \begin{subfigure}[b]{0.19\textwidth}
    \centering
        \includegraphics[scale=0.118]{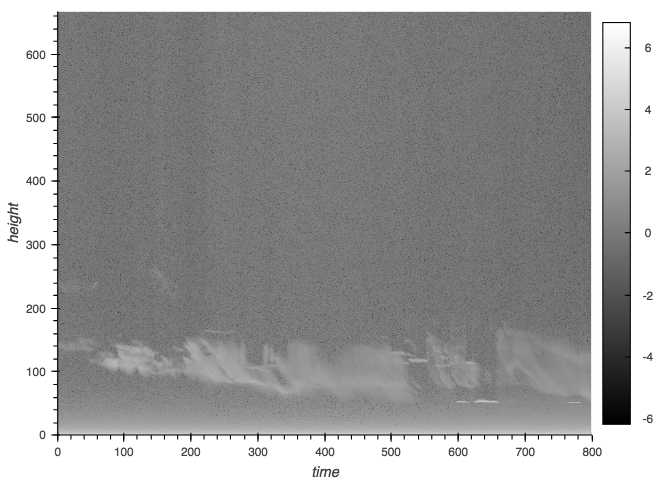}
    \end{subfigure}
    \begin{subfigure}[b]{0.19\textwidth}
    \centering
        \includegraphics[scale=0.118]{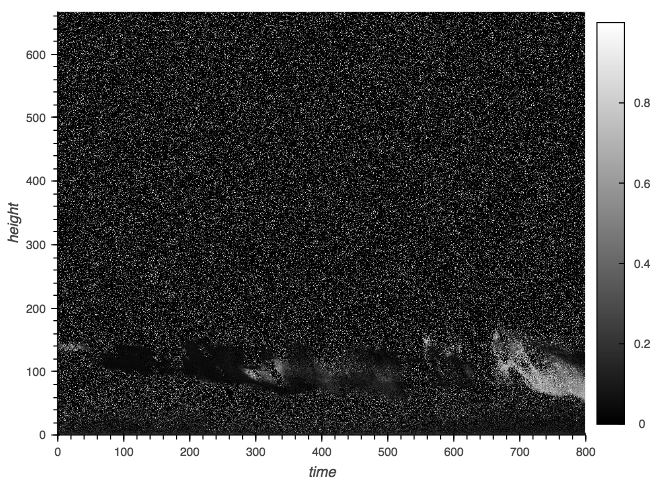}
    \end{subfigure}
    \begin{subfigure}[b]{0.19\textwidth}
    \centering
        \includegraphics[scale=0.118]{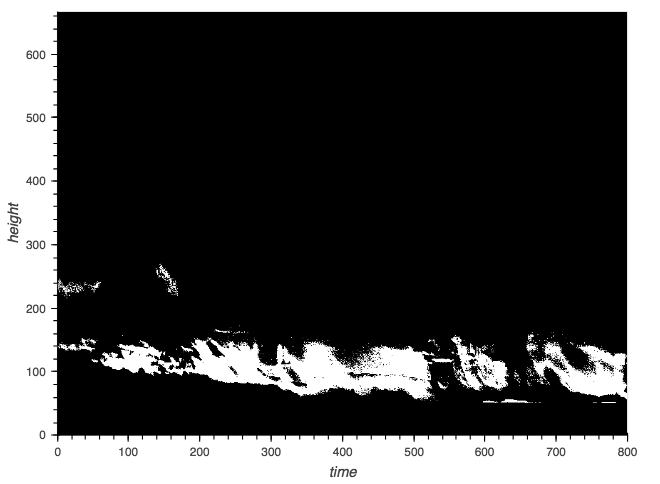}
    \end{subfigure}
    \begin{subfigure}[b]{0.19\textwidth}
    \centering
        \includegraphics[scale=0.118]{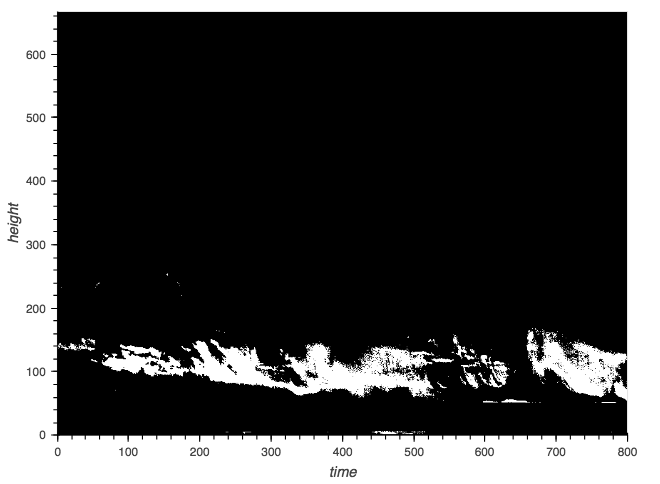}
    \end{subfigure}
    \begin{subfigure}[b]{0.19\textwidth}
    \centering
        \includegraphics[scale=0.118]{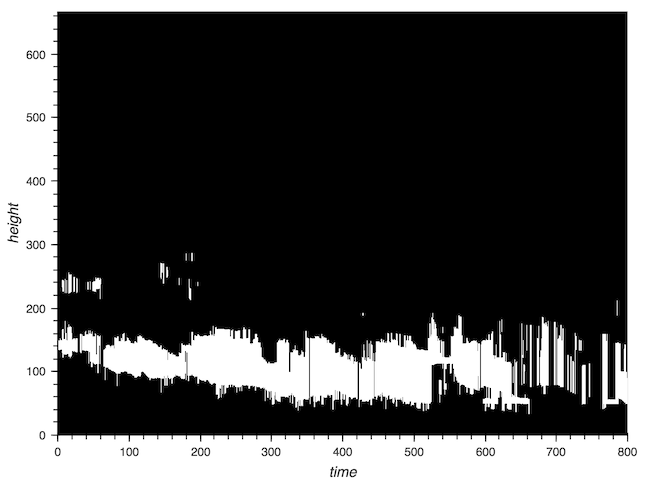}
    \end{subfigure}
\caption{Model results for MPL data at the Oliktok site:
         Top row is May 11th, 2016 (bins 2080 to 2880),
         bottom row is May 13, 2015 (bins 0 to 800).
         \textbf{(a)}: MPL backscatter profile.
         \textbf{(b)}: MPL linear depolarization ratio.
         \textbf{(c)}: hand-labeled cloud mask.
         \textbf{(d)}: segmentation model output.
         \textbf{(e)}: MPLCMASK cloud mask.}
\label{fig:oliktok_figures}
\end{figure*}

%% file: conclusion.tex
We successfully use train a FCN to segment clouds from lidar imagery.
We show our semi-supervised training method outperformed 
the MPLCMASK data product
and verify it improves the overall performance of the model.
Likewise, we analyze model performance on 
lidar data from different observations sites (mid-latitude versus polar)
and potential transfer learning to make the model more robust.
Initial results are promising, but requires further investigation.
In the future, we want to investigate if data seasonality (winter vs. summer)
impacts model training and cloud detection.
Another area of interest is applying our model to satellite lidar data
from the Cloud-Aerosol Lidar and Infrared Pathfinder Satellite Observation (CALIPSO) satellite
and to lidar data from High Spectral Resolution Lidar (HSRL) systems.
Furthermore, we are interested in if adding another input data channel, 
specifically a ceilomoter lidar data, improves model performance.

%% file: acknowledgements.tex
The research described in this work is part of the 
Deep Science for Scientific Discovery Initiative at Pacific Northwest National Laboratory. 
It was conducted under the Laboratory Directed Research and Development Program at PNNL, 
a multiprogram national laboratory operated by Battelle for the U.S. Department of Energy.
This research was performed using PNNL Institutional Computing at Pacific Northwest National Laboratory.
Also, we would like to thank the Pacific Northwest National Laboratory 2017 Quickstarter Initiative 
for initial funding for this work.
Additionally, 
we would like to thank the Atmospheric Radiation Measurement (ARM) Data Center 
for the 30smplcmask1zwang data stream data.